\documentclass[lettersize,journal]{IEEEtran}
\usepackage{amsmath,amsfonts}
\usepackage{amssymb}
\usepackage{algorithm}
\usepackage{algpseudocode}
\usepackage{array}
\usepackage[caption=false,font=normalsize,labelfont=sf,textfont=sf]{subfig}
\usepackage{textcomp}
\usepackage{stfloats}
\usepackage{url}
\usepackage{verbatim}
\usepackage{graphicx}
\usepackage{cite}
\usepackage{csvsimple}
\usepackage{booktabs}
\usepackage{float}
\usepackage[table]{xcolor}
\begin{document}
	\title{Clustering-based Transfer Learning for Dynamic Multimodal MultiObjective Evolutionary Algorithm}
	
	\author{Li Yan, \IEEEmembership{Member, IEEE,}
		Bolun Liu,
		Chao Li,
		Jing Liang, \IEEEmembership{Senior Member, IEEE,} 
		Kunjie Yu, \IEEEmembership{Member, IEEE,} 
		Caitong Yue, \IEEEmembership{Member, IEEE,} 
		Xuzhao Chai,
		and Boyang Qu, \IEEEmembership{Member, IEEE}
		\thanks{This work was supported by the National Natural Science Foundation of China (Grant Nos. 62576372, 62373389), Leading talents of science and technology in the Central Plain of China (254200510055), the Key Research and development projects of Henan Province (Grant No. 241111210100), the Science and Technology Innovation talents of Colleges and Universities in Henan Province (Grant No. 24HASTIT037), the Graduate Education Reform Project of Henan Province (Grant Nos. 2023SJGLX188Y), the Postgraduate Education Reform and Quality Improvement Project of Henan Province (YJS2026YBGZZ20), Frontier Exploration Projects of Longmen Laboratory (LMQYTSKT031).}
		\thanks{Li Yan, Bolun Liu, Chao Li, Xuzhao Chai, and Boyang Qu are with the School of Automation and Electrical Engineering, Zhongyuan University of Technology, Zhengzhou 450007, China (email: yanli@zut.edu.cn; lbl202511@163.com; lichaoedu@126.com; xzchai@zut.edu.cn; quboyang@zut.edu.cn).}
		\thanks{Jing Liang, Kunjie Yu, and Caitong Yue are with the School of Electrical Engineering, Zhengzhou University, Zhengzhou 450001, China (e-mail: liangjing@zzu.edu.cn; yukunjie@zzu.edu.cn; zzuyuecaitong@163.com).}}
	
	\maketitle
	
	\begin{abstract}Dynamic multimodal multiobjective optimization presents the dual challenge of simultaneously tracking multiple equivalent pareto optimal sets and maintaining population diversity in time-varying environments. However, existing dynamic multiobjective evolutionary algorithms often neglect solution modality, whereas static multimodal multiobjective evolutionary algorithms lack adaptability to dynamic changes. To address above challenge, this paper makes two primary contributions. First, we introduce a new benchmark suite of dynamic multimodal multiobjective test functions constructed by fusing the properties of both dynamic and multimodal optimization to establish a rigorous evaluation platform. Second, we propose a novel algorithm centered on a Clustering-based Autoencoder prediction dynamic response mechanism, which utilizes an autoencoder model to process matched clusters to generate a highly diverse initial population. Furthermore, to balance the algorithm's convergence and diversity, we integrate an adaptive niching strategy into the static optimizer. Empirical analysis on 12 instances of dynamic multimodal multiobjective test functions reveals that, compared with several state-of-the-art dynamic multiobjective evolutionary algorithms and multimodal multiobjective evolutionary algorithms, our algorithm not only preserves population diversity more effectively in the decision space but also achieves superior convergence in the objective space.
			
	\end{abstract}
	

	\section{Introduction}
	\IEEEPARstart{O}{VER} the past few years, Dynamic Multi-objective Optimization Problems (DMOPs) have garnered significant attention \cite{ref1, ref2, ref3, ref4, ref5} due to their inherent complexity and widespread applications in domains such as shop scheduling \cite{ref6},  resource allocation \cite{ref7}, and dynamic priority scheduling \cite{ref8}. The core characteristic of DMOPs is that their objective functions or constraints change over time, causing the pareto optimal front (POF) and pareto optimal set (POS) to vary accordingly. Which poses a formidable challenge for optimization algorithms, they must not only converge to the current POF but also continuously track environmental changes and rapidly re-converge to the new POF. To tackle the challenges of dynamic multi-objective optimization, researchers have developed a series of efficient evolutionary algorithms revolving around the core ideas of diversity enhancement, memory, prediction, and knowledge transfer. Furthermore, to evaluate their performance, researchers have designed a series of standard test functions, such as DF \cite{ref9}, FDA \cite{ref10} and JY \cite{ref11}.However, while current research has made significant strides in enhancing objective-space performance, the issue of multiple POSs in the decision space has been largely overlooked.
	\begin{figure}[!t]
		\centering
		\includegraphics[scale=0.1]{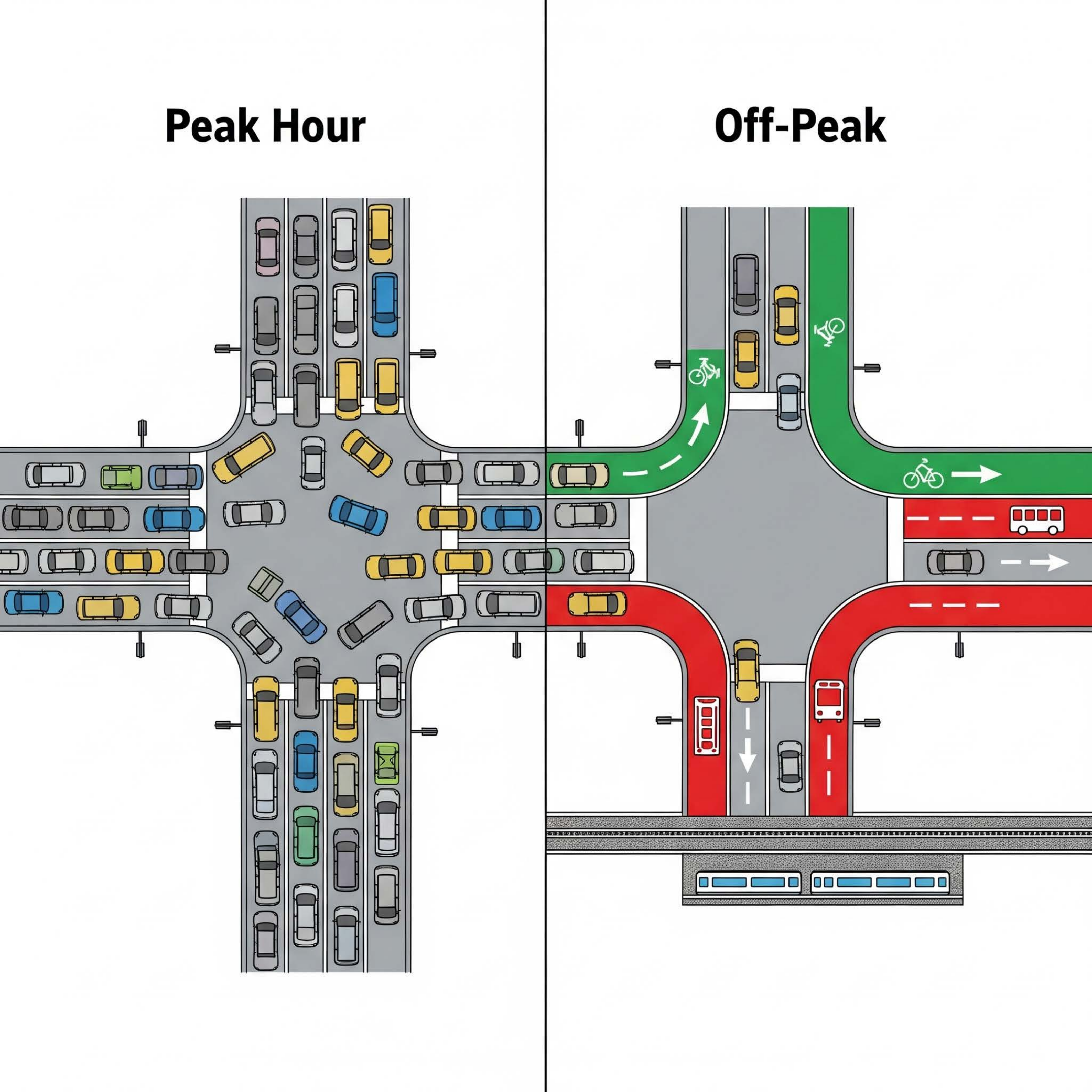}
		\caption{ Illustration of the dynamic multimodal multiobjective urban traffic problem}
		\label{fig_2}
	\end{figure}
	
	In the field of multi-objective optimization, there exists another class of challenging problems known as multimodal multiobjective optimization problems (MMOPs), which are characterized by the presence of multiple POSs in the decision space \cite{ref12}. Unlike traditional DMOPs that only require finding a single POS, MMOPs demand that algorithms maintain a high degree of diversity in the decision space to discover all coexisting POSs in their entirety. However, most existing algorithms are designed for static environments, once the environment undergoes dynamic changes, they struggle to adapt quickly and continuously track all the updated POSs.
	
	When the challenges of dynamism and multi-modality intertwine, they form the even more complex dynamic multimodal multiobjective optimization problems (DMMOPs) \cite{ref13}. In DMMOPs, a dynamically evolving POF may correspond to two or more different POSs at the same point in time. Although finding any single POS may suffice for certain applications, providing multiple POSs offers the decision-maker a richer set of superior choices \cite{ref14, ref15}. Therefore, researching algorithms capable of completely discovering all dynamic POSs is a crucial direction for advancing the field.
	
	Urban traffic and road optimization is a typical DMMOP \cite{ref16}. Its complexity is first reflected in multi-objectivity, as decision-makers must make trade-offs among multiple conflicting goals such as travel efficiency, congestion alleviation, and environmental protection. Secondly, this trade-off process is highly dynamic, stemming from real-time changes in travel demand and traffic conditions. For example, as illustrated in Figure 1, the optimal travel choice between a congested ``peak hour'' and a smooth ``off-peak'' period is distinctly different, requiring the system to make rapid strategic adjustments. Finally, multi-modality provides diverse means to achieve this dynamic trade-off but also complicates the problem: the same travel objective can be met through various modes like cars, buses, subways, or bicycles, which form a vast and complex space of strategic combinations due to their different energy efficiencies and capacities.
	
	DMMOPs present a significant challenge due to the intricate interplay of dynamism, multi-objectivity, and multi-modality. This combination imposes stringent demands on multiobjective evolutionary algorithms (MOEAs) to balance convergence and diversity across both decision and objective spaces in dynamic environments. The central aim in solving DMMOPs is to achieve efficient tracking and comprehensive identification of all equivalent POSs upon an environmental change. However, existing approaches are fundamentally ill-equipped for this task. Dynamic multiobjective evolutionary algorithms (DMOEAs) are mainly focused on convergence, which allows them to rapidly identify the new POF. However, their lack of diversity-enhancing strategies poses a significant challenge when tackling DMMOPs, making it difficult to locate all POSs. Conversely, multimodal multiobjective evolutionary algorithms (MMOEAs) excel at preserving population diversity to find multiple POSs in static environments. Yet, they typically lack the dynamic response mechanisms required to adapt to changes. As a result, when the optimal solutions shift, MMOEAs face difficulties in rapid adaptation, causing a dramatic drop in their convergence performance.
	
	It is worth noting that simply combining existing dynamic handling techniques with multimodal optimization methods cannot effectively solve DMMOPs. This is because the former emphasizes convergence while the latter emphasizes diversity, which can conflict with each other. Therefore, to address this challenge, it has become a pressing task to research and develop integrally designed dynamic multimodal multi-objective optimization algorithms (DMMOEAs) that are specifically tailored to the characteristics of DMMOPs.
	
	Despite the prevalence of DMMOPs in the real world, related research remains insufficient, particularly with a lack of standard benchmarks for comprehensively evaluating algorithm performance. To address above challenge, this paper designs and proposes a novel set of dynamic multi-modal multi-objective test functions, named DMMFs. This test suite integrates diverse time-varying mechanisms to construct dynamically evolving POF and multiple POSs. These mechanisms endow the test functions with complex dynamic characteristics, such as discreteness and deformation. Its creation fills a current void in the availability of test functions for this domain.
	
	To effectively solve DMMOPs, this paper proposes a new algorithm denoted as Clustering-based Autoencoder and adaptive niching (CAE-AN). The key to this algorithm lies in its novel dynamic response strategy, which effectively balances the diversity and convergence of the predicted population by partitioning the multiple solution sets in the dynamic environment. Furthermore, CAE-AN enhances its ability to maintain diversity throughout the optimization process by incorporating an adaptive niching strategy into its static optimizer.Specifically, the main contributions of this paper
	are clarified as follows:
	
	\begin{enumerate}
		\item To construct test instances for DMMOPs that reflect the characteristics of real-world applications, this paper introduces a comprehensive design framework. This framework enables the generation of three distinct classes of DMMOP test functions, each embodying different challenging properties. Leveraging this framework, we have designed and present a suite of 12 specific test functions, collectively designated as DMMFs, to provide a standardized benchmark for the rigorous performance evaluation and fair comparison of relevant algorithms.
		
		\item To solve DMMOPs, this paper proposes a novel algorithm named CAE-AN, where the proposed algorithm first uses DBSCAN to cluster historical solutions, then an Autoencoder (AE) is used to predict the new positions of the solution clusters, and seeds the new population with these predicted solutions for rapid adaptation. Furthermore, an adaptive niching technique is also integrated to enhance diversity.
		
		\item To validate the practical performance of the proposed algorithm, we designed a series of comparative experiments to evaluate its effectiveness in solving DMMOPs. Several state-of-the-art DMOEAs and MMOEAs were selected as comparison algorithms, where these algorithms cannot directly handle the coexisting dynamic and multi-modal characteristics in DMMOPs, we modified and extended them to establish fair and competitive baselines. 
	\end{enumerate}
	
	The remainder of this paper is organized as follows. Section II first reviews the related literature and clarifies the motivation for this study. Following this, Section III and Section IV are dedicated to detailing the newly constructed DMMF benchmark suite and the proposed CAE-AN algorithm, respectively. Section V then presents a comprehensive experimental study to validate the performance and effectiveness of CAE-AN. Finally, Section VI concludes the paper and outlines future research directions.
	
	\section{RELATED WORK AND MOTIVATION}
	The design objective of DMMOEAs is to address multimodal multiobjective optimization problems in dynamic environments. This necessitates algorithms capable of maintaining a well-distributed set of solutions in both objective and decision spaces, while rapidly converging to all Pareto optimal solution sets after an environmental change. However, research on DMMOEAs is still in its nascent stages, and designing an efficient algorithm that simultaneously handles both multimodality and dynamism remains a significant challenge. Considering the substantial progress made in DMOEAs and MMOEAs in recent years, these advancements provide a valuable foundation for DMMOEA research. Therefore, the following section will briefly review the relevant work in DMOEAs and MMOEAs.
	\subsection{Multimodal Multiobjective Optimization}
	1) Pareto-dominance-based algorithms: To find all Pareto optimal sets in MMOPs, researchers have proposed a series of algorithms based on the fundamental framework of Pareto dominance. As an early representative, Omni-optimizer \cite{ref16} pioneered the combination of crowding distance in both decision and objective spaces with non-dominated sorting, laying the foundation for maintaining population diversity. Building on this, subsequent research has seen the emergence of more diverse niching strategies to enhance exploration capabilities, such as the dual-niching technique used in DNEA \cite{ref17}, the ring topology utilized by MO\_Ring\_PSO\_SCD \cite{ref18}, and the clustering method applied in MMOEA/DC \cite{ref19}.
	To tackle more complex challenges, particularly the conflict between maintaining diversity and accelerating convergence, researchers have designed specialized mechanisms. For instance, TriMOEA-TA\&R \cite{ref20}, proposed by Liu et al., employs a dual-archive strategy to collaboratively balance the two, while their subsequent work, CPDEA \cite{ref21}, resolves the imbalance problem during the search process through a convergence-penalized technique. Furthermore, other innovative perspectives have appeared in the field, such as the grid-guided search algorithm proposed by Qu et al.~\cite{ref22} and an MMOEA that focuses solely on knee point solutions.
	\par 2) Decomposition-based algorithms: The core idea of decomposition-based MMOEAs is to transform the original problem into multiple single-objective subproblems for solving, by using a decomposition technique \cite{ref11}. In early explorations, Hu et al.~\cite{ref23} proposed assigning a fixed number ($k$) of solutions to each subproblem and performing individual selection based on a fitness indicator using the Penalty-Based Boundary Intersection (PBI) method \cite{ref12}. However, setting the parameter $k$ appropriately is extremely difficult. To resolve this issue, Tanabe et al.~\cite{ref24} cleverly replaced $k$ with an insensitive neighborhood size parameter in their algorithm, MOEA/D-AD.
	As research progressed, more sophisticated decomposition strategies were proposed. For example, MOEA/D-MM, proposed by Peng and Ishibuchi \cite{ref25}, combines a subpopulation framework with the decomposition strategy to tackle large-scale MMOPs. Tanabe et al.~\cite{ref26} introduced a dynamic update mechanism in MOEA/D-ADA, which adjusts the population size of each subproblem in real-time using ``Addition'' or ``Deletion'' operators. Furthermore, Paland and Bandyopadhyay \cite{ref27} took a different approach by employing Graph Laplacian-based clustering to decompose the decision space while simultaneously using reference vectors to decompose the objective space, thereby implementing a novel dual-space decomposition strategy.
	\par 3) Indicator-based algorithms: The core of Indicator-Based MMOEAs is to guide the population's selection process by defining a specific fitness indicator. However, the calculation of such indicators often relies on sensitive parameters, leading to challenges similar to those faced by decomposition-based algorithms.
	To address this issue, researchers have conducted a series of explorations. For example, Li et al.~\cite{ref28} proposed MMEA-WI, an algorithm that designs a specialized weight fitness indicator to evaluate the quality of solutions. Building on this, their subsequent work, HREA \cite{ref29}, further developed a hierarchy ranking method to effectively handle complex problems with local POFs. Additionally, NIMMO, proposed by Tanabe et al.~\cite{ref30}, also belongs to this category; it uses carefully selected performance indicators to guide the population's search efficiently within different niches.
	\subsection{Dynamic Multiobjective Optimization}
	1) Diversity-based DMOEAs: Maintaining population diversity is crucial for tracking optima in DMOPs subject to environmental changes. Some algorithms address potential diversity loss directly. Deb et al. \cite{ref31}, for example, modified NSGA-II\cite{ref32} to create DNSGA-II-A and DNSGA-II-B. DNSGA-II-A injects random individuals to preserve diversity, whereas DNSGA-II-B employs a higher mutation rate on the existing population for the same purpose. While measures like random individual introduction or increased mutation can enhance adaptability, they carry a potential trade-off. Overemphasizing diversity generation may compromise the convergence speed, particularly in complex problem landscapes, by detracting from efficient exploitation of promising search regions.
	
	Post-change response strategies, such as reinitialization or strong variation injection, aim to restore population diversity following environmental alterations. However, this reactive  approach demonstrates limited efficiency when applied to DMMOPs. Its inherent drawbacks involve the potential loss of historical evolutionary information and the challenge, given constrained computational resources, of effectively addressing the simultaneous tracking and optimization requirements of multiple, potentially dispersed Pareto fronts exhibiting diverse dynamics.
	\par 2) Memory-based DMOEAs: Memory-based approaches are specifically designed to solve DMOPs characterized by periodic environmental changes. Their core mechanism involves storing high-quality solutions from past environments and, upon detecting similarity between the current and a historical environment, transferring and reusing these solutions to accelerate the algorithm's convergence. For example, the hybrid strategy HMPS proposed by Liang et al.\cite{ref33} first identifies the similarity between environments and then applies distinct strategies based on its existence.
	
	Admittedly, this approach performs excellently when dealing with periodic problems, but its true technical bottleneck lies in two key challenges: how to accurately measure the similarity between environments, and how to efficiently extract valuable knowledge from vast and complex historical data.
	\par 3) Prediction-based DMOEAs: Prediction-based Methods represent an active area of research within the field of DMOPs. Their core idea is to extract patterns of change by analyzing historical data and to use these patterns to predict the future POS. Owing to their superior performance in tracking dynamic optimal solutions, these methods have garnered widespread attention from researchers \cite{ref34,ref35,ref36,ref37,ref38,ref39,ref40}.
	
	A variety of implementation paths exist. For example, Li et al. \cite{ref34} proposed a multi-strategy adaptive selection method employing a non-inductive transfer learning paradigm, which aims to precisely match the most effective knowledge to the current environment. Yu et al. \cite{ref35}, designed a correlation-guided layered prediction strategy that divides the population into three collaborative groups and tailors distinct prediction tasks for each. Building on this, Yu et al. \cite{ref36} further constructed a historical learning-based framework to assist static optimizers in more comprehensively drawing information from historical data.
	
	\subsection{Dynamic Multimodal Single-objective Optimization}
	\par 1)Memory-based Methods: Memory-based dynamic response strategies aim to guide the current search by storing and reusing information from historical environments. The underlying principle is that optimal solutions from the past may remain valid or be close to new optimal solutions in the new environment. Taking the research by Luo et al.~\cite{ref41} as an example, this method maintains elite antibodies from previous environments in a memory archive. Combined with the random initialization of some individuals, it strikes a balance between ``inheritance'' and ``innovation''---it not only preserves valuable optimization experience but also ensures population diversity by introducing new individuals, thus preventing premature convergence of the algorithm.
	
	The main advantage of such strategies lies in their efficiency when dealing with periodic or gradual environmental changes. However, their limitations are also quite apparent: when the patterns of environmental change are complex, drastic, or completely unknown, historical information may become obsolete or even misleading, leading to a decline in algorithm performance.
	\par 2)Prediction-based Methods: Unlike memory-based strategies that rely on historical solutions, prediction-based strategies look to the future, proactively constructing models to anticipate the new positions of optimal solutions after an environmental change. Following this technical path, researchers have progressively evolved from direct prediction to more complex adaptive prediction methods.
	
	Initial explorations, such as the work by Wu et al.~\cite{ref42}, directly used Support Vector Machines (SVM) for regression prediction of peak positions in the new environment. Building on this, to achieve a more targeted response, Yan et al.~\cite{ref43} proposed a ``divide and conquer'' hybrid model strategy. This approach first partitions niches using an SVM classifier and then applies separate autoregressive predictions to each independent sub-population. To tackle more complex dynamic environments, Ahrari et al.~\cite{ref44} designed a more advanced adaptive multi-level prediction strategy. This method not only improves accuracy by incorporating knowledge from a longer historical window but also introduces an adaptive mechanism to dynamically select the optimal prediction level, significantly enhancing the robustness of the prediction and the precision of the optimization search.

	\subsection{Motivations}
	DMMOPs represent a significant and challenging class of problems. Their complexity stems from the superposition of three core characteristics: first, the problem's objectives or constraints change dynamically over time; second, multiple distinct optimal solution regions exist at any given time, exhibiting multi-modal properties; and third, the algorithm must balance multiple conflicting objectives, which constitutes its multi-objective nature. This unique combination requires an algorithm to possess exceptional flexibility and precision to simultaneously track multiple solution sets that evolve in different ways within a changing environment.
	
	However, existing DMOEAs generally prove inadequate when dealing with DMMOPs, with their bottlenecks manifesting in two main areas. The first is inadequate prediction capability; conventional models struggle to capture the complex, non-linear trajectories of the optimal solution sets, leading to inaccurate predictions and failing to provide effective guidance for the algorithm's subsequent search. The second is a poor diversity maintenance strategy; when multiple optimal solution sets evolve differently, existing algorithms lack differentiated management mechanisms. Their response strategy of uniformly optimizing all solution sets at once makes it difficult to effectively maintain population diversity and can easily lead to the loss of some solution sets.
	
	To precisely address the aforementioned difficulties, this paper proposes a novel algorithm named CAE-AN. This algorithm directly addresses the deficiencies of existing methods, with its core being a two-stage strategy of identification followed by prediction. First, the algorithm utilizes clustering techniques to accurately identify multiple independent POSs from the current population. Subsequently, for each identified set, an AE model is employed to learn its respective historical evolutionary data to independently predict its future dynamics. Furthermore, to better balance convergence and diversity during the optimization process, CAE-AN also introduces an adaptive niching radius operation. This ensures that while efficiently tracking multiple targets, the diversity of solutions is effectively maintained, preventing the algorithm from premature convergence.
	\begin{table*}[!t]  
		\centering
		\caption{DMMF Test Problems and Their Properties}
		\label{tab:dmmf_properties}
		\begin{tabular}{ccccccc}
			\toprule
			Testproblem name & objectives & POF geometry & POS geometry & dynamic & \begin{tabular}[c]{@{}c@{}}scalable number\\ of Pareto set\end{tabular} & \begin{tabular}[c]{@{}c@{}}Coexistence of\\ global and local Pareto set\end{tabular} \\
			\midrule
			DMMF1  & 2 & concave & Linear & \begin{tabular}[c]{@{}c@{}}static POF,\\ dynamic PS\end{tabular} & $\times$ & $\times$ \\
			DMMF2  & 2 & convexity-concavity & convexity-concavity & \begin{tabular}[c]{@{}c@{}}dynamic POF and\\ POS\end{tabular} & $\times$ & $\times$ \\
			DMMF3  & 2 & convexity-concavity & concave & \begin{tabular}[c]{@{}c@{}}dynamic POF and\\ POS\end{tabular} & $\times$ & $\times$ \\
			DMMF4  & 2 & convexity-concavity & static-dynamic & \begin{tabular}[c]{@{}c@{}}dynamic POF and\\ POS\end{tabular} & $\times$ & $\times$ \\
			DMMF5  & 2 & convexity-concavity & sine wave & \begin{tabular}[c]{@{}c@{}}dynamic POF,\\ static POS\end{tabular} & $\times$ & $\times$ \\
			DMMF6  & 3 & convexity-concavity & convexity-concavity & \begin{tabular}[c]{@{}c@{}}dynamic POF and\\ dynamic POS\end{tabular} & $\times$ & $\times$ \\
			DMMF7  & 2 & concave & sine wave & \begin{tabular}[c]{@{}c@{}}static POF,\\ dynamic POS\end{tabular} & $\surd$ & $\times$ \\
			DMMF8  & 2 & convexity-concavity & linear-convex & \begin{tabular}[c]{@{}c@{}}dynamic POF and\\ POS\end{tabular} & $\times$ & $\times$ \\
			DMMF9  & 2 & concave & convexity-concavity & \begin{tabular}[c]{@{}c@{}}dynamic POF and\\ POS\end{tabular} & $\times$ & $\surd$ \\
			DMMF10 & 2 & convexity-concavity & sine wave & \begin{tabular}[c]{@{}c@{}}dynamic POF and\\ POS\end{tabular} & $\times$ & $\times$ \\
			DMMF11 & 2 & convex & sine wave & \begin{tabular}[c]{@{}c@{}}static POF,\\dynamic POS\end{tabular} & $\times$ & $\times$ \\
			DMMF12 & 2 & linear-concavity & sine wave & \begin{tabular}[c]{@{}c@{}}dynamic POF and\\ POS\end{tabular} & $\times$ & $\times$ \\
			\bottomrule
		\end{tabular}
	\end{table*}
	
	\begin{figure*}[!t]
		\centering
		\subfloat[]{\includegraphics[width=3.5in]{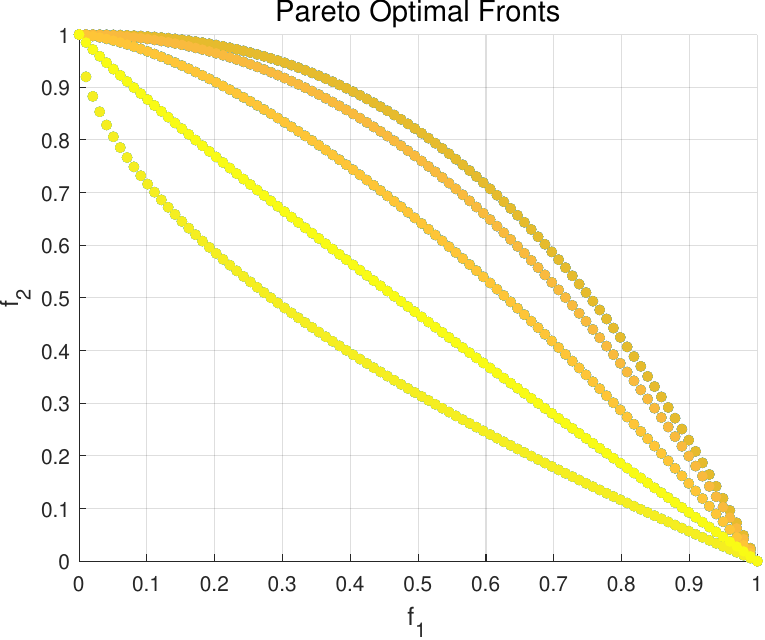}%
			\label{fig_first_case}}
		\hfil
		\subfloat[]{\includegraphics[width=3.5in]{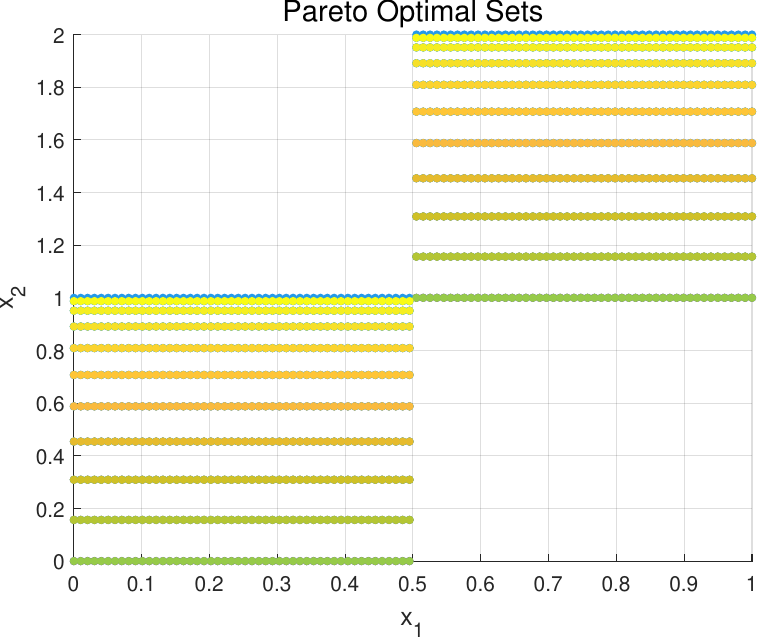}%
			\label{fig_second_case}}
		\caption{ An example of DMMF. (a) Illustration of the dynamic POF in the objective space. (b) Illustration of the dynamic POS in the decision space.}
		\label{fig_4}
	\end{figure*}
	\section{DYNAMIC MULTIMODAL MULTIOBJECTIVE FUNCTION}
	Numerous and highly challenging DMMOPs are prevalent in real-world applications. Nevertheless, the availability of test function suites that can effectively model such problems is, to date, quite restricted. To foster further scientific inquiry in this field, this paper develops a new set of dynamic multi-modal multi-objective optimization test functions, designated as DMMFs, which are derived from the properties of practical problems. The design principles, specific instances, and key properties of the DMMFs are presented in this section.
	\subsection{ Definition of DMMF}
	 Definition of Pareto Domination:For a minimization problem, given two solutions x and y, the solution x Pareto dominates the solution y, denoted x $\prec$ y, if solution x and solution y satisfies,
	 \begin{equation}
	 	\begin{split}
	 		&\forall i \in \{1, 2, \dots, M\}, f_i(x) \leq f(y), \text{ and} \\
	 		&\exists i \in \{1, 2, \dots, M\}, f_i(x) < f(y).
	 	\end{split}
	 \end{equation}
	 In equation (2), M represents the dimension of the objective space. For the given solutions x and y, if solution x does not satisfy the condition for Pareto domination of solution y, and solution y does not satisfy the condition for Pareto domination of solution x, then solution x and solution y do not dominate each other, denoted x $\prec$ y and y $\prec$ x.
	 
	 An optimization problem is formally defined as a DMMOP if and only if it simultaneously satisfies the following two core characteristics:
	 
	 \begin{enumerate}
	 	\item \textbf{Multimodality:}
	 	
	 	At any given time instant $t$, the problem satisfies at least one of the following conditions:
	 	\begin{enumerate}
	 		\item The existence of at least one local POS.
	 		
	 		\item The existence of at least two equivalent global POSs that map to the same point on the Pareto front in the objective space but possess different vector representations in the decision space.
	 	\end{enumerate}
	 	
	 	\item \textbf{Dynamics:}
	 	
	 	The problem exhibits explicit or implicit time-varying characteristics in its objective functions, constraints, or environmental parameters. This time-varying nature induces at least one of the following evolutionary patterns:
	 	\begin{enumerate}
	 		\item Structural transformations of the POF, including modifications to its geometry, spatial position, or solution density distribution.
	 		
	 		\item Dynamic evolution of local/global POSs, manifested as changes in their quantity, spatial configuration, or equivalence relationships, such as the emergence, disappearance, bifurcation, or coalescence of solutions.
	 		
	 		\item Transitional changes in the modality characteristics of solutions, particularly the inter-conversion between local and global Pareto-optimal states over time.
	 	\end{enumerate}
	 \end{enumerate}
	 
	 \subsection{ Design Principle of DMMFs}
	  To comprehensively evaluate the performance of DMMOAs, this study constructs a benchmark suite of test functions exhibiting both dynamic and multi-modal characteristics. This test suite aims to simulate complex, real-world optimization scenarios in two key aspects: 
	  \begin{enumerate}
	  	\item \textbf{Dynamic Characteristics:} The test functions incorporate time-varying features, including dynamic changes in the objective functions, constraints, and POS over time. These changes emulate common real-world phenomena such as environmental disturbances, data stream updates, and system parameter drift, thereby assessing the algorithms' dynamic adaptation capabilities.	  	
	  	\item \textbf{Multi-modal Characteristics:} The test functions include multiple global and local optima, simulating complex multi-modal optimization problems. This design assesses an algorithm's ability to concurrently search for, capture, and maintain multiple high-quality solutions within a complex solution space, i.e., its diversity preserving capabilities.
	  \end{enumerate} 
	  \par By integrating these two characteristics, the constructed test suite provides a robust evaluation foundation for developing novel DMMOEA with enhanced practicality and generalization ability.The proposed framework structure is as follows:
	  \begin{equation}
	  	\left\{
	  	\begin{aligned}
	  		& \min \mathbf{F}(\mathbf{x}, \mathbf{y}, t) = (f_1, f_2, \dots, f_m)^T \\
	  		& \text{where } f_j = h_j(\mathbf{x}, \mathbf{G}(t)) \cdot [1 + g(\mathbf{x}, \mathbf{y}, \mathbf{G}(t))]\\
	  		& \quad\quad\mathbf{x} = (x_1, x_2, \dots, x_p)^T \\
	  		& \quad\quad\mathbf{y} = (y_1, y_2, \dots, y_q)^T \\
	  		& \quad\quad j = 1, 2, \dots, m \\ 
	  		& \text{subject to } g(\mathbf{x}, \mathbf{y}, \mathbf{G}(t)) = 0
	  	\end{aligned}
	  	\right.
	  \end{equation}
	   
	  Within this framework, $m$ represents the number of objective functions; $\mathbf{x} \in \mathbb{R}^p$ and $\mathbf{y} \in \mathbb{R}^q$ are the $p$-dimensional and $q$-dimensional decision variables, respectively; $\mathbf{G}(t)$ is a time-varying parameter function that introduces dynamic characteristics; and $\mathbf{F}(\mathbf{x}, \mathbf{y}, t)$ is the vector containing the $m$ objective functions to be optimized.
	  
	  The ingenuity of this framework's design is that the function $\mathbf{G}(t)$ is responsible for controlling the dynamic changes of the POF and the POS, whereas the structure of the decision variables, $(\mathbf{x}, \mathbf{y})$, determines their geometry at any given time instant.
	  
	  Based on this modular design framework, test functions with diverse dynamic and multi-modal characteristics can be flexibly constructed to facilitate a comprehensive evaluation of algorithmic performance.
	  
	  Here, a simple example is provided to illustrate the construction method for these test functions. The first step is to construct a dynamic multi-objective function and determine its corresponding Pareto front. Let the function h() be defined by the following equation:
	  \begin{equation}
	  		h(x, G(t)) = 1-x_1^{G(t)}
	  \end{equation}
	 
	  When the function $g(x, G(t))$ reaches its minimum value of 0, the Pareto front for the problem can be determined by equations (3) and (4).
	  \begin{equation}
	  	\begin{cases}
	  		f_2=1-f_1^G \\
	  		0 \le f_1, f_2 \le 1
	  	\end{cases}
	  \end{equation}
	  
	  The second step of the construction process involves designing the function $g(x, G(t))$ to build multiple dynamic POSs. To achieve this, we first define a specific time-dependent function; for instance, let the time-varying parameter be $G(t) = \sin(0.5\pi t)$. Based on this setting, the function $g(x, G(t))$ is expressed as follows:

	  \begin{equation}
	  	g(x, G(t)) = 
	  	\begin{cases} 
	  		(x_2 - G(t))^2 & \text{if } x_1 < 0.5 \\
	  		(x_2 - (G(t) + 1))^2 & \text{if } x_1 \geq 0.5 
	  	\end{cases}
	  	\label{eq:g_function_example}
	  \end{equation}
	  
	  In the current design, the condition $g=0$ defines two disjoint solution paths: when $x_1 < 0.5$, the optimal solution is $x_2 = G(t)$; when $x_1 \geq 0.5$, the optimal solution is $x_2 = G(t) + 1$. This generates two distinct and time-varying POSs, both of which map to the same dynamic POF, as illustrated in Figure~2.
	  
	  In essence, by synergistically combining the $h$ function, which controls the POF dynamics, with a strategically designed $g$ function for the POS topology, this framework provides a powerful and flexible method for generating complex and realistic benchmark problems.
	  
	  \begin{figure*}[!t]
	  	\centering
	  	\subfloat[Type I]{\includegraphics[width=0.32\textwidth]{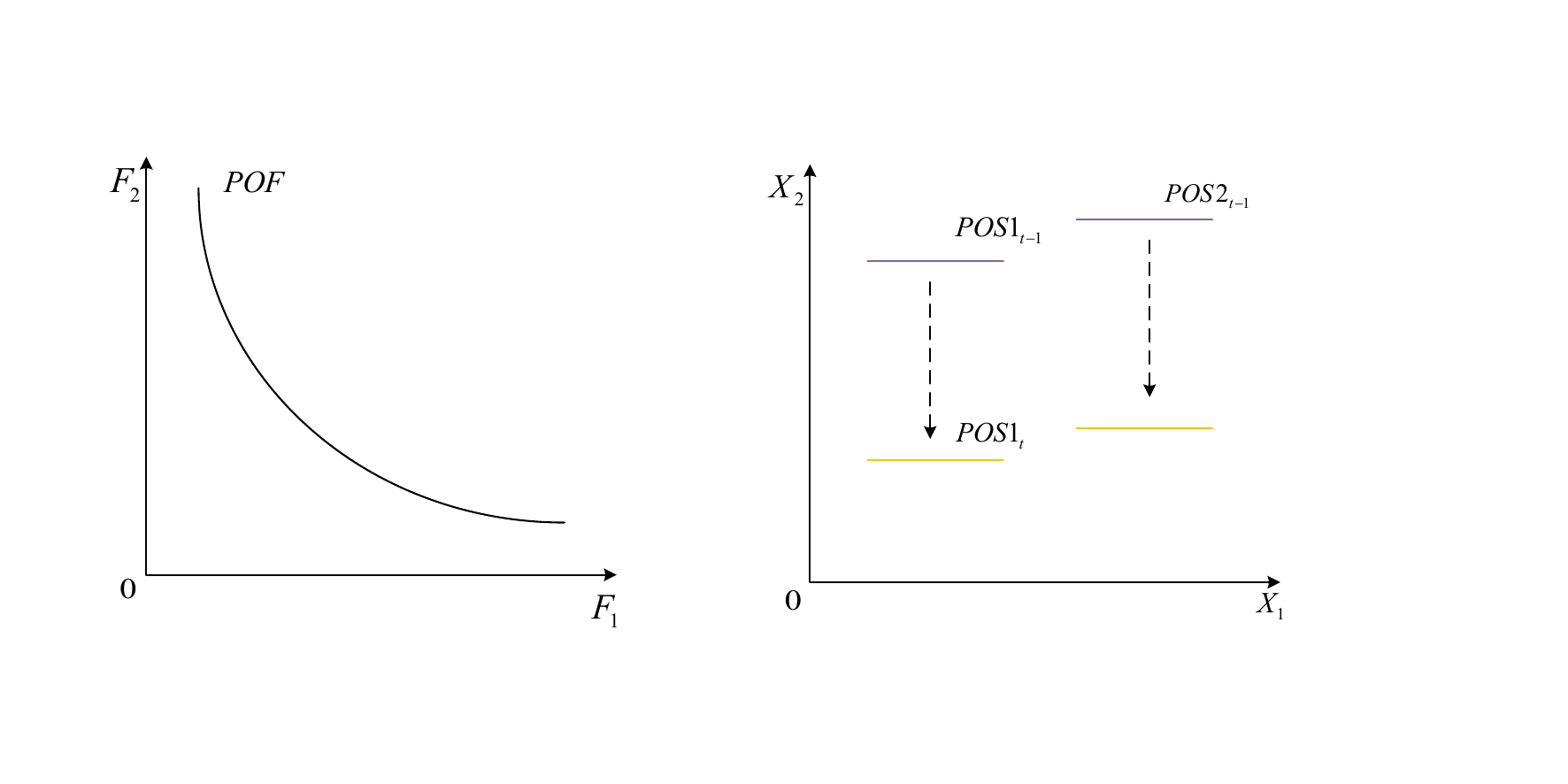}%
	  		\label{fig_first_case}}
	  	\hfill 
	  	\subfloat[Type II]{\includegraphics[width=0.32\textwidth]{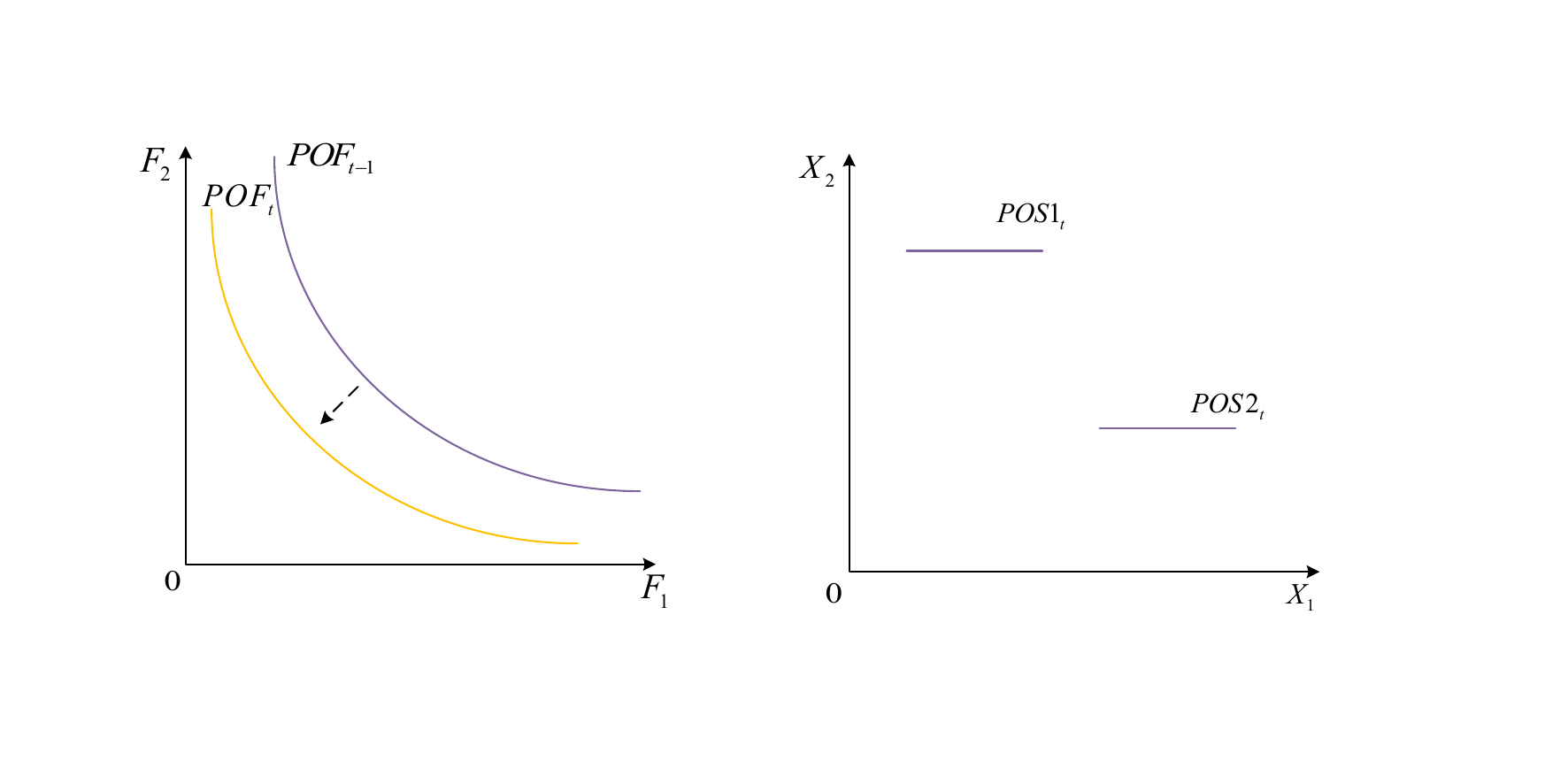}%
	  		\label{fig_second_case}}
	  	\hfill 
	  	\subfloat[Type III]{\includegraphics[width=0.32\textwidth]{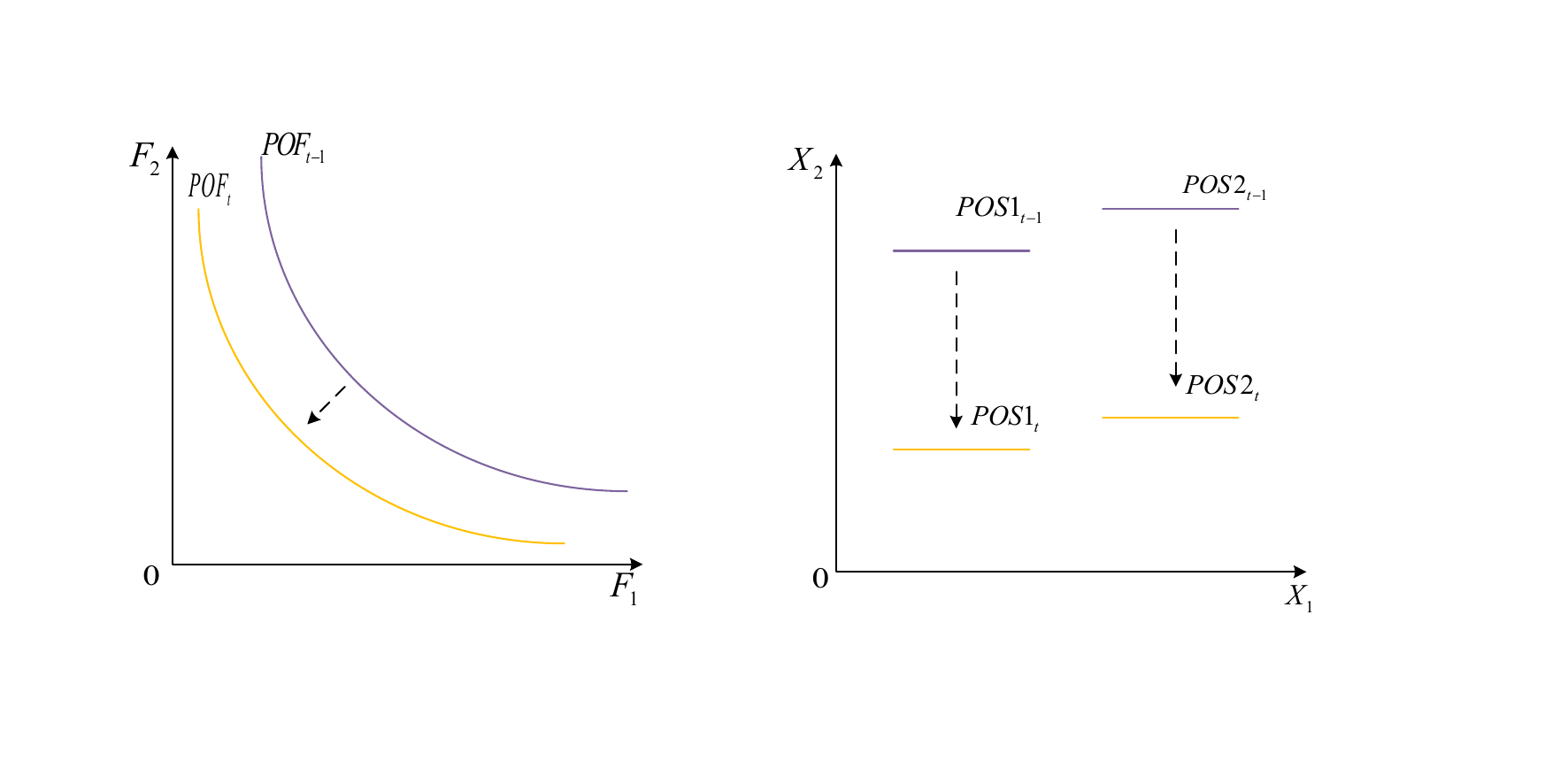}% 
	  		\label{fig_third_case}}
	  	\caption{Visualization of the dynamic multi-modal multi-objective test function types: (a) Type I structure; (b) Type II structure; (c) Type III structure.}
	  	\label{fig_5} 
	  \end{figure*}
	 \subsection{Instances and Characteristics of DMMF}
	 To comprehensively address the dual challenges of environmental dynamism and solution space multimodality, we propose a tripartite classification scheme for the DMMF. As illustrated in Figure~3, the framework is categorized into three distinct types based on the temporal characteristics of its POF and POS:
	 	\begin{enumerate}
	 		\item \textbf{Type I: The POSs change over time while the POF remains stationary.} This configuration maintains a stationary POF while allowing dynamic evolution of the PS across environmental changes. The dynamic POS manifests in two distinct subtypes:
	 		\begin{enumerate}
	 			\item Subtype I-A: Fully dynamic POS, where all optimal solutions undergo continuous spatial reconfiguration in response to environmental shifts.
	 			\item Subtype I-B: Partially dynamic POS, characterized by a hybrid architecture where specific solution clusters maintain spatial stability while others exhibit adaptive relocation.
	 		\end{enumerate}
	 		The inherent heterogeneity in solution space dynamics introduces varying optimization complexity levels, particularly evident in Subtype I-B where static solution clusters present lower search difficulty compared to their dynamic counterparts. This structural dichotomy is visually substantiated in Figure~3(a)-(c).
	 		
	 		\item \textbf{Type II: The POF changes over time while the POS remains stationary.} Contrasting with Type I, this category features a temporally evolving PF paired with a stationary PS configuration. The environmental dynamics exclusively influence the objective space geometry, while the solution space maintains fixed topological characteristics. Consequently, optimization efforts encounter uniform search difficulty across all PS regions, as the invariant solution space preserves consistent landscape properties throughout the optimization process. This configuration proves particularly relevant for problems requiring stable solution maintenance under shifting performance criteria.
	 		
	 		\item \textbf{Type III: Both POF and POSs change over time.} This type of problem is the most complex, characterized by the concurrent dynamic changes in both the POF and the POS. This dual dynamism requires the algorithm to be able to simultaneously adapt to variations in both the objective space and the decision space.
	 		
	 \end{enumerate}
	 \par Existing MOEAs face significant challenges in effectively solving multi-modal problems in dynamic environments, which involves finding and maintaining multiple Dynamic Pareto Optimal Sets (DPOSs). The core challenges primarily stem from the following points:
	 
	 \begin{itemize}
	 	\item The time-varying nature of objective functions and constraints greatly increases the difficulty for algorithms to completely and continuously track the entire Dynamic Pareto Optimal Front (DPOF) in the objective space.
	 	
	 	\item Significant differences in convergence difficulty among various DPOSs can cause an algorithm to become easily trapped in local optima. That is, it may be drawn towards easier-to-converge solution sets, thereby failing to discover the full set of globally distributed optimal solutions.
	 	
	 	\item Most DMOEAs primarily focus on performance in the objective space, while paying insufficient attention to the diversity and distribution of solutions in the decision space. This directly hinders their ability to discover and preserve all equivalent Pareto solution sets.
	 	
	 	\item Mainstream MMOEAs are typically designed for static problems and lack effective mechanisms for detecting and responding to environmental changes. Consequently, they tend to respond slowly after an environment update, failing to adapt quickly to the new distribution of optimal solutions.
	 \end{itemize}
	 
	 This classification framework establishes fundamental theoretical distinctions in the field of dynamic multi-modal optimization, thereby providing systematic guidance for algorithm selection and performance benchmarking. Its progressive complexity gradient from Type I to Type III reflects the escalating computational demands of simultaneously handling multimodality and dynamism, which is a critical consideration for real-world adaptive optimization systems. Based on the guiding principles of this framework, we have designed a series of 12 test functions, the detailed characteristics of which are summarized in Table I.

	\section{THE PROPOSED METHOD}
	This section elaborates on the proposed DMMOEA, CAE-AN. The central idea of this algorithm is that when an environmental change occurs, its key dynamic response strategy is activated. This strategy employs DBSCAN to identify multiple POSs and then uses an AE to independently predict these identified POSs. Concurrently, its static optimization part integrates an adaptive niche strategy into NSGA-II to improve the distribution of the obtained POSs.
	\par The remainder of this section is organized as follows: Section IV.A describes the general framework of CAE-AN, and Section IV.B provides a detailed elaboration of the CAE strategy's implementation.

\begin{algorithm}[!t]
	\caption{CAE-AN} \label{alg:pan_dmmoea_corrected_unified}
	\begin{algorithmic}[1] % Line numbering enabled
		\Statex \textbf{Input:} DMMFs: $F_t(x)$, Population size: $N$, Radius in DBSCAN: $\epsilon$, Minimum points in DBSCAN: $\eta$ 
		\Statex \textbf{Output:}The set for saving POS in different moments: $\mathbf{POS}$
		\State $\mathbf{POS} \gets \emptyset$
		\State $t \gets 0$
		\State $\mathbf{initPop} \gets$ Generate $N$ random solutions
		\While{stopping criterion is not met}
		\State $\mathbf{POS} \gets \Call{MMOEA}{F_0(x), \mathbf{initPop}, N}$
		\State $\mathbf{POS} = \mathbf{POS} \cup \mathbf{POS}_t$
		\If{environment changes}
		\State $t \gets t + 1$
		\If{$t < 2$}
		\State $\mathbf{initPop} \gets Reinitialize a population$
		\Else
		\State $\mathbf{initPop} \gets \Call{CAE}{\mathbf{POS}, t, N, \epsilon, \eta}$
		\EndIf
		\EndIf
		\EndWhile
		
		\State \Return $\mathbf{POS}$
	\end{algorithmic}
\end{algorithm}
	\subsection{The General Framework}
	We propose the CAE-AN framework to address the challenges of DMMOPs. The framework extends traditional dynamic algorithms with a key innovation in its dynamic response mechanism. It consists of two main components: a static optimizer and a dynamic predictive model. The static optimizer is a modified NSGA-II that incorporates an adaptive niching strategy to maintain solution diversity. The dynamic component is designed to generate a high-quality initial population for each new environment. It does so by training an AE model on clustered and matched historical POSs to predict the solution distribution after a change. This process provides the static optimizer with an initial population that is well distributed and forward-looking, which in turn initiates the search process.
	
	Algorithm~1 presents the detailed pseudocode of CAE-AN, which is designed to efficiently solve DMMOPs. During the initialization phase (lines 1--3), the algorithm establishes an archive to store the POSs from all generations and creates an initial working population. The main loop of the algorithm (lines 4--15) employs distinct population initialization strategies depending on the current time step $t$,
	During the startup phase ($t < 2$), to facilitate a cold start and initial exploration, the population is randomly initialized at $t=0$ (lines 10). 
	During the steady-state phase ($t \ge 2$), the framework activates its core dynamic response mechanism. In this stage, the algorithm retrieves the POS archives from the two preceding time steps, $t-2$ and $t-1$. It then applies the CAE strategy (line 12) to generate a high-quality initial population that possesses both excellent distribution and proactive, predictive characteristics. This population is subsequently injected into the static optimization algorithm to commence the search process for the current time step.
	
	Notably, to enhance the diversity maintenance capability of the static optimizer, we have improved the classic NSGA-II algorithm by incorporating an adaptive niching technique. The core innovation of this technique lies in an adaptive radius adjustment mechanism: we designed a novel formula that can dynamically adjust the niche radius based on the iteration process and the individual density within the niche, thereby maintaining population diversity more flexibly. The specific formula is as follows:
	\begin{equation}
		R_i(g) = R_0 \cdot \left( 1 - \alpha \cdot \frac{g}{G_{\max}} \right) \cdot \left( 1 + \alpha \cdot Var_i \right)
	\end{equation}
	
	Here, $R_i(g)$ is the new radius for individual $i$ at generation $g$, $Var_i$ is the fitness variance within its niche, and the term with $g/Gmax$ ensures the search space gradually narrows as the algorithm approaches its final generation.
	
	\begin{figure*}[!t]
		\centering
		\includegraphics[scale=1]{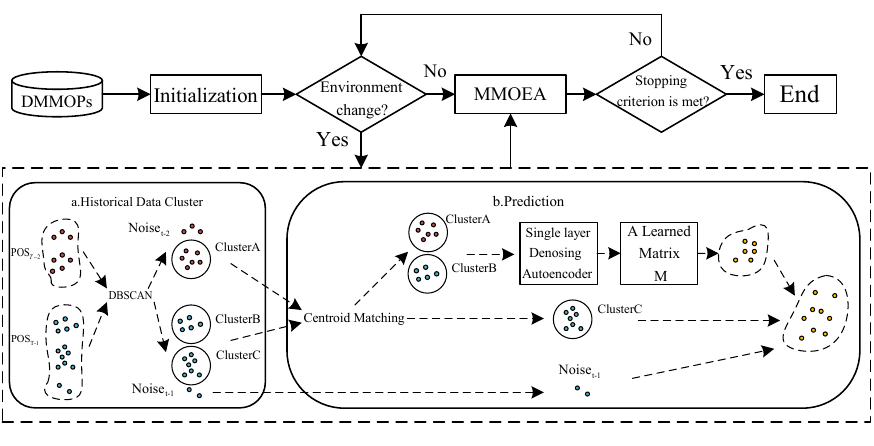}
		\caption{ Flowchart of CAE Strategy using DBSCAN Clustering and AE Prediction.}
		\label{fig_3}
	\end{figure*}
	
\begin{algorithm}[!t]
	\caption{CAE} \label{alg:dbc_ae_strategy_unified}
	\begin{algorithmic}[1]
		\Statex \textbf{Input:}$\mathbf{POS}$: Pareto Optimal Sets: POS, time instance:$t$,  Population size:$N$, Radius in DBSCAN: $\epsilon$, Minimum points in DBSCAN: $\eta$
		\Statex \textbf{Output:}$\mathbf{initPop}$: Initial pop. for cycle $t$
		\State $\mathbf{Clus}_{t-2}\gets \Call{DBSCAN}{\mathbf{POS}_{t-2}, \epsilon, \eta}$
		\State $\mathbf{Clus}_{t-1}\gets \Call{DBSCAN}{\mathbf{POS}_{t-1}, \epsilon, \eta}$
		\State $\mathbf{Cents}_{t-2} \gets \Call{CalcCents}{\mathbf{Clus}_{t-2}}$
		\State $\mathbf{Cents}_{t-1} \gets \Call{CalcCents}{\mathbf{Clus}_{t-1}}$
		\State $\mathbf{MatchedPairs} \gets \Call{FindMatches}{\mathbf{Cents}_{t-2}, \mathbf{Cents}_{t-1}}$
		\State $\mathbf{PredictedSols} \gets \emptyset$
		\ForAll{cluster pair \textbf{in} $\mathbf{MatchedCluster}$}
		\State $\mathcal{M} \gets \Call{AE\_Learn}{\mathbf{C}_{t-2} \to \mathbf{C}_{t-1}}$
		\State $\mathbf{NewSols} \gets \Call{AE\_Predict}{\mathbf{C}_{t-1}, \mathcal{M}}$
		\State $\mathbf{PredictedSols} \gets \mathbf{PredictedSols} \cup \mathbf{NewSols}$
		\EndFor
		\State $\mathbf{initPop} \gets \mathbf{PredictedSols}$
		\State \Return $\mathbf{initPop}$
	\end{algorithmic}
\end{algorithm}
	
	\subsection{CAE Strategy}
	This section details the proposed DMMOP strategy, termed CAE. While many existing DMOEAs have successfully incorporated knowledge transfer mechanisms, including notable approaches utilizing AEs for trend prediction, a common limitation is the insufficient consideration of scenarios where multiple, structurally diverse, and independently evolving POSs coexist within the dynamic environment. Addressing above challenge, the core innovation of the CAE strategy lies in its two-stage approach: firstly, DBSCAN is employed to cluster non-dominated solutions within the current population, thereby effectively identifying and partitioning distinct POSs. Secondly, for each independently identified POS, a dedicated AE is subsequently trained using its historical information to perform fine-grained dynamic prediction. This 'cluster-first, then predict-independently' mechanism enables the prediction models to better adapt to the unique evolutionary characteristics of disparate POSs.
	
	The procedural implementation of this strategy follows a clear sequence of steps, with its core objective being the prediction of dynamically changing POSs.
	
	The entire process begins with historical information archiving (Lines 1--2), where the algorithm archives the non-dominated solution sets from two preceding consecutive time steps, $t-2$ and $t-1$. This data serves as the foundation for subsequent analytical tasks.
	
	Subsequently, the algorithm enters the historical POS identification and partitioning phase (Lines 3--4). Here, the DBSCAN clustering algorithm is independently applied to the non-dominated solution sets from time steps $t-2$ and $t-1$. The choice of DBSCAN is motivated by its effectiveness in identifying arbitrarily shaped clusters and its robustness to noise, which is crucial for handling diverse and complex POS structures. Through this clustering process, the multiple POSs within each historical snapshot are clearly identified and partitioned.
	
	Next, the algorithm performs POS feature extraction and matching (Lines 5--6). First, the centroid of each identified POS cluster is computed to serve as its compact, representative feature in the objective space. Then, by calculating and comparing the Euclidean distances between these centroids, clusters from time step $t-2$ are paired with their closest counterparts from time step $t-1$. This step successfully establishes the temporal correspondence of an evolving POS, laying the groundwork for independent prediction and tracking.
	
	Finally, for each successfully matched POS, the algorithm executes the core step of independent predictive model construction and inference (Lines 7--11). To achieve efficient dynamic tracking, this strategy does not employ a complex non-linear AE. Instead, it constructs a dedicated linear mapping model for each POS, designed to capture the primary dynamics of its evolution between time steps with high computational efficiency.
	
	Specifically, the model aims to solve for a transformation matrix $\mathbf{M}$ that best describes the linear relationship from a historical POS state, $\mathbf{X}_{t-1}$, to the current state, $\mathbf{X}_t$. This matrix is not learned through iterative training but is obtained directly via an analytical solution to a regularized least-squares problem . The formula is as follows:
	\begin{equation}
		\mathbf{M} = (\mathbf{X}_t \mathbf{X}_{t-1}^T) (\mathbf{X}_{t-1} \mathbf{X}_{t-1}^T + \lambda\mathbf{I})^{-1}
	\end{equation}
	where $\lambda$ is the regularization parameter and $\mathbf{I}$ is the identity matrix.
	
	Once the matrix $\mathbf{M}$ is computed, it is used to extrapolate and predict the state of the current POS ($\mathbf{X}_t$) at the next time step, $t+1$, denoted as $\hat{\mathbf{X}}_{t+1}$. This prediction process combines two key components. First, the core dynamics are applied by performing the matrix multiplication $\mathbf{M}\mathbf{X}_t$, which represents the learned linear transformation. Second, a perturbation term, $v$, calculated from the model's recent Mean Squared Error (MSE), is added to estimate the stochastic dynamics. This term accounts for the system's random drift or other minor dynamics not fully captured by the linear model.
	The final predicted solution is generated by the formula:
	\begin{equation}
		\hat{\mathbf{X}}_{t+1} = \mathbf{M}\mathbf{X}_t + v
	\end{equation}
	
	In summary, through this procedural workflow of ``cluster, match, and predict'', the strategy establishes a dedicated, lightweight predictive model for each evolving POS. This ``divide and conquer'' approach enables it to address multi-modality in a targeted manner, thereby capturing the inherent and often heterogeneous complex dynamics within DMMOPs with greater precision and robustness.
	
	\section{EXPERIMENTAL RESULTS}
	
	\subsection{Performance Indicators} 
	 To evaluate the effectiveness of the algorithms, this paper employs two performance indicators: mean inverted generational distance (MIGD) and mean decision space inverted generational distance (MIGDx). Their definitions are as follows:
	 
	 \textit{1)Mean Inverted Generational Distance:}The Inverted Generational Distance (IGD) is a key performance metric used to comprehensively evaluate an algorithm's convergence and diversity. It is calculated by taking the average distance from the points on the True POF to their nearest neighbors in the approximated POF generated by the algorithm. Therefore, a lower IGD value signifies that the algorithm's solution set is closer to the optimal front in both convergence and diversity, resulting in better overall performance [46]. The metric is defined as follows:
	 \begin{equation}
	 	IGD(POF^*, POF) = \frac{\sum_{p^* \in POF^*} \min_{p \in POF} \| p^* - p \|}{|POF^*|}
	 \end{equation}
	 
	 Here, $POF^*$ refers to the True Pareto Front, whereas $POF$ represents the approximated front generated by the algorithm.
	 
	 In the context of DMMOPs, the constantly shifting environment implies that an IGD value from a single computation offers merely a ``snapshot'' of performance, which is inadequate for fully evaluating an algorithm's ability to adapt dynamically. For this reason, the MIGD metric was introduced. By averaging the IGD scores measured over several successive environmental changes, it provides a more comprehensive, long-term evaluation of an algorithm's dynamic performance [8].
	 
	 MIGD is mathematically defined as follows:
	 \begin{equation}
	 	MIGD(POF_t^*, POF_t) = \frac{\sum_{t \in T} IGD(POF_t^*, POF_t)}{|T|}
	 \end{equation}
	 
	 where $T$ is the set of discrete timepoints in a run, and $|T|$ is its cardinality.
	 
	 \textit{2)Mean Inverted Generational Distance in decision space:}While the conventional \text{IGD} metric is dedicated to assessing algorithmic performance in the \textit{objective space}, it fails to gauge performance within the \textit{decision space}. For multimodal optimization problems, obtaining a diverse POS in the decision space is just as crucial as locating the POF in the objective space.
	 
	 To address above challenge, the \text{IGDx} metric was developed based on the principles of \text{IGD}, tailored specifically for evaluating solution set quality in the decision space. Analogous to \text{IGD}'s computation of distances between POF, \text{IGDx} quantifies the convergence and diversity in the decision space by computing the average distance between the true $POS^*$ and the approximated set $POS$ produced by the algorithm.
	 
	 Consequently, a lower \text{IGDx} score signifies that the algorithm's solutions achieve a better approximation of the true optimal set and a superior distribution in the decision space, which in turn serves as evidence of the algorithm's enhanced multimodal search capabilities~[46]. This metric is mathematically defined as:
	 \begin{equation}
	 	IGDx(POS^*, POS) = \frac{\sum_{p^* \in POS^*} \min_{p \in POS} \| p^* - p \|}{|POS^*|}
	 \end{equation}
	 
	 MIGDx is mathematically defined as follows:
	 \begin{equation}
	 	MIGDx(POS_t^*, POS_t) = \frac{\sum_{t \in T} IGD(POS_t^*, POS_t)}{|T|}
	 \end{equation}
	 
	 \subsection{Benchmark Problems} 
	 \begin{table}[H]
	 	\centering 
	 	\caption{CONFIGURATIONS OF DYNAMIC CHANGES ON THE DMMFS} 
	 	
	 	\begin{tabular}{|c|c|c|c|c|}
	 		\hline
	 		Configuration & C$_1$ & C$_2$ & C$_3$ & C$_4$ \\
	 		\hline
	 		$n_t$ & 5 & 10 & 5 & 10 \\
	 		\hline
	 		$\tau_t$ & 10 & 5 & 5 & 10 \\
	 		\hline
	 	\end{tabular}
	 	
	 \end{table}
	 Following the discussion in Section III, a suite of twelve DMMOP test functions is employed to evaluate algorithm performance. These test functions integrate characteristics of both prevalent dynamic multi-objective test functions and multi-modal multi-objective test functions. In dynamic multi-objective test functions, multiple environment settings are typically used to represent different types of environmental changes. The primary temporal parameter controlling these dynamic changes t is calculated as:
	 $t = \frac{1}{n_t} \left\lfloor \frac{\tau}{\tau_t} \right\rfloor$.
	 where $n_t$ represents the frequency of environmental change, $\tau_t$ denotes the severity of the environmental change, and $\tau$ represents the current generation number.  To 
	 comprehensively assess algorithm performance under diverse environmental conditions, each test function in this study is configured with four distinct environment settings. These settings are detailed in Table II.
	 
	\subsection{Compared Algorithms}
	For a comprehensive performance evaluation of the proposed CAE-AN algorithm, we selected six state-of-the-art algorithms for a comparative study, drawn from two distinct categories: DMOEA and MMOEA.
	
	The selected DMOEAs include D-NSGA-II\cite{ref31}, a baseline approach that randomly re-initializes 20\% of the population upon environmental change; DIP-DMOEA\cite{ref38}, a prediction-driven algorithm that utilizes an Artificial Neural Network (ANN) to identify new optimal regions; and Tr-DMOEA\cite{ref45}, which leverages knowledge transfer to address environmental dynamics.
	
	The MMOEAs chosen for comparison are DN-NSGA-II\cite{ref46}, which employs a niching technique in the decision space to promote diversity; MMEA-WI\cite{ref28}, which selects solutions based on a consolidated distance metric in both objective and decision spaces; and MO-RING-SCD-PSO\cite{ref18}, which integrates niching with a special crowding distance calculation.
	
	Notably, standard MMOEAs lack intrinsic mechanisms for dynamic adaptation. To establish a fair basis for comparison, we endowed these three MMOEAs with the same dynamic response capability as our proposed method. This was achieved by introducing an AE model to predict the location of new optima, thereby allowing them to effectively respond to environmental changes.
	
	\subsection{Parameter Setting}
	To ensure a fair comparison, the parameters for the compared algorithms were set according to the recommendations in their respective publications \cite{ref18,ref28,ref31,ref38,ref45,ref46}.
	The population size was set to 100 for two-objective DMMOPs and 150 for three-objective DMMOPs. All algorithms employed Simulated Binary Crossover (SBX) and Polynomial Mutation (PM) for crossover and mutation operations, respectively. The maximum number of generations was set to 30 × $\tau$, corresponding to 30 environmental changes.
	
	Both the proposed CAE-AN and the compared algorithms were executed independently 20 times on each test function.  All implementations and experiments were conducted using MATLAB 2022a.
	\subsection{ Experimental Results and Analysis}
	\par In terms of performance evaluation in the objective space, Table III presents the comparative MIGD metric values for each algorithm (detailed results for all comparison algorithms are provided in Table I of the supplementary material due to space limitations). Statistical analysis based on the Wilcoxon rank-sum test reveals that the advantage of the CAE-AN algorithm is highly significant, as it achieved the best performance on 39 out of the 48 test instances. In contrast, the other six comparison algorithms achieved the best results on only one to three instances each. This result strongly demonstrates that CAE-AN possesses superior convergence capabilities in the objective space of the DMMF test suite. This superiority is primarily attributed to its unique core mechanism: after an environmental change, CAE-AN first performs clustering analysis and matching on solutions from the two preceding time steps, then utilizes the successfully matched pairs to train an AE model. This model learns the patterns of environmental change to predict potentially superior solution regions in the new environment, which are then refined by the optimizer, thereby leveraging historical information to achieve rapid convergence.
	
	\begin{table}[!t]
		\centering
		\caption{STATISTICS OF PERFORMANCE COMPARISONS OF CAE-AN AND SIX ALGORITHMS ACCORDING TO MIGD AND MIGDx INDICATORS.}
		\label{tab:performance_comparison_vertical_2}
		
		% Table for MIGD
		\begin{tabular}{|l|c|}
			\hline
			\multicolumn{2}{|c|}{\textbf{MIGD}} \\
			\hline
			\textbf{Comparison} & \textbf{+/-/=} \\
			\hline
			CAE-AN vs D-NSGA-II & 36/12/0 \\
			CAE-AN vs Tr-DMOEA & 27/18/3 \\
			CAE-AN vs DIP-DMOEA & 39/9/0 \\
			CAE-AN vs D-DN-NSGA-II & 40/8/0 \\
			CAE-AN vs D-MMEA-WI & 44/4/0 \\
			CAE-AN vs D-MO-RING-SCD-PSO & 42/5/1\\
			\hline
		\end{tabular}
		
		\vspace{1em} 
		
		\begin{tabular}{|l|c|}
			\hline
			\multicolumn{2}{|c|}{\textbf{MIGDx}} \\
			\hline
			\textbf{Comparison} & \textbf{+/-/=} \\
			\hline
			CAE-AN vs D-NSGA-II & 34/13/1 \\
			CAE-AN vs Tr-DMOEA & 30/15/3 \\
			CAE-AN vs DIP-DMOEA & 38/10/0 \\
			CAE-AN vs D-DN-NSGA-II & 30/15/3 \\
			CAE-AN vs D-MMEA-WI & 44/4/0 \\
			CAE-AN vs D-MO-RING-SCD-PSO & 33/14/1\\
			\hline
		\end{tabular}
		
		\begin{flushleft}
			\small
			“+”, “-”, and “=” denote CAE-AN performs significantly worse than, better than, and similar to the compared algorithm based on the Wilcoxon’s rank sum test at 0.05 significant level.
		\end{flushleft}
	\end{table}
	Regarding performance evaluation in the decision space, Table III reports the MIGDx metric(the detailed data are shown in Table II in the supplementary file), which reflects the quality of the solution distribution. The statistical results once again confirm the leading position of CAE-AN, which secured the best MIGDx values on 38 instances, while all comparison algorithms combined were superior on only the remaining 10. This indicates that CAE-AN can more effectively generate and maintain a high-quality, well-distributed set of solutions in the decision space. This performance advantage is primarily attributed to the aforementioned CAE strategy, which helps to generate well-distributed and promising candidate solutions in the new environment, thereby comprehensively enhancing its performance in the decision space.
	
	To provide further intuitive evidence of the dynamic process and qualitative results, Figures 5 and 6 illustrate the algorithm's performance from different perspectives. Figure 5 depicts the IGD values over time for each algorithm on 12 representative test functions. It can be clearly observed that at the vast majority of time points, CAE-AN's IGD curve is significantly lower than those of the other algorithms. This not only proves the speed and efficiency of its dynamic response mechanism but also corroborates that its optimization capability within each static environment is also outstanding. Figure 6, by comparing the POSs found by each algorithm against the true POS on the DMMF12 test function, intuitively reveals their diversity maintenance capabilities. The figure shows that the POS found by CAE-AN is highly consistent with the true POS in both shape and distribution. In contrast, Tr-DMOEA and DIP-DMOEA only find fragments of the POS, as their mechanisms neglect diversity in the decision space. Meanwhile, other algorithms like D-NSGA-II, despite potentially finding a more complete POS initially, ultimately lose parts of the solution set after continuous environmental changes because they struggle to cope with the simultaneous dynamic evolution of multiple POSs. These visual results provide strong corroborating evidence for the comprehensive performance advantage of CAE-AN.
	
	\begin{figure*}[!t]
		\centering
		\includegraphics[scale=0.2]{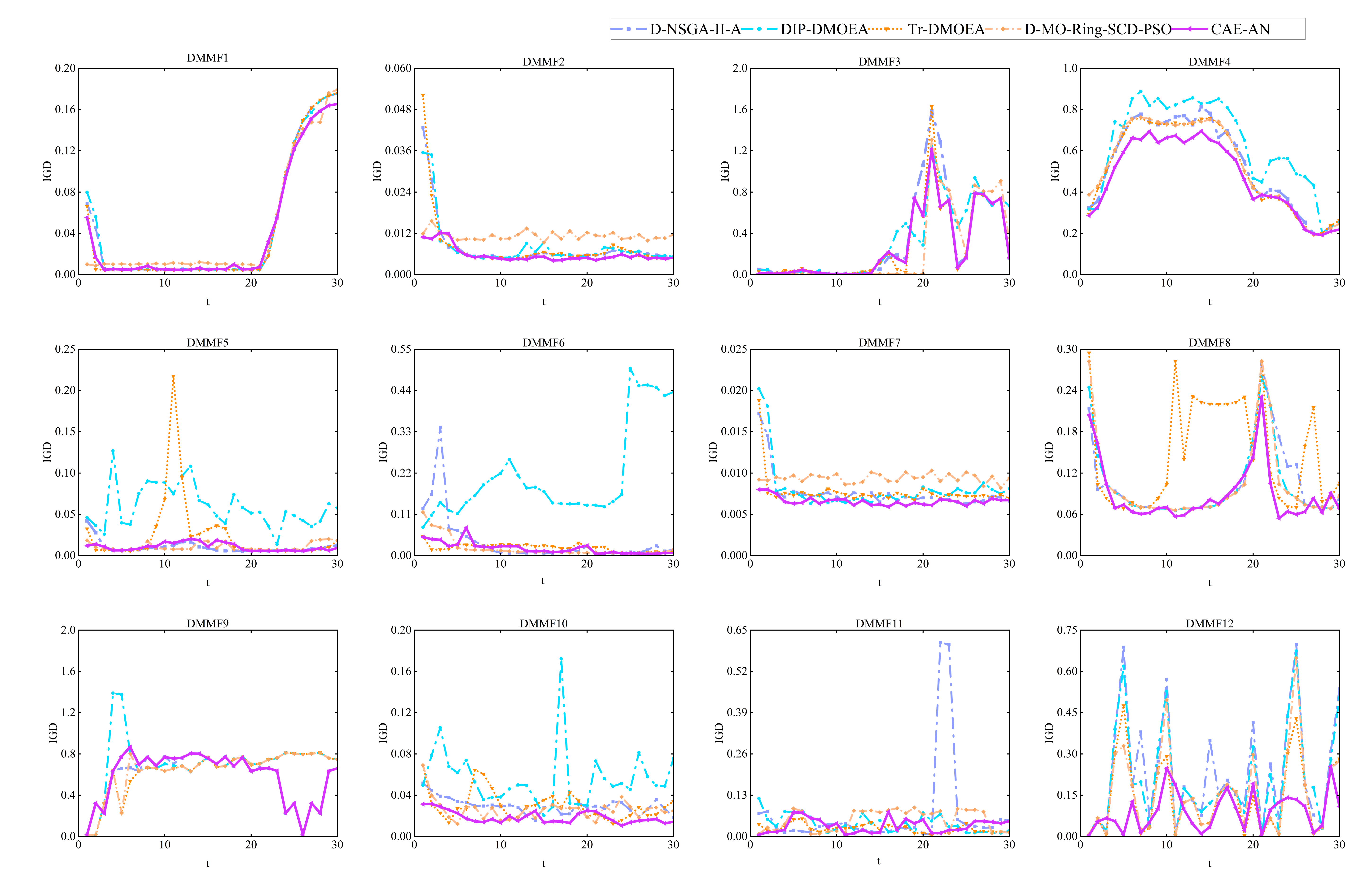}
		\caption{IGD Results of Different Algorithms on the DMMF1–DMMF12 Test Functions}
		\label{fig_7}
	\end{figure*}
	
	\begin{figure*}[!t]
		\centering
		\subfloat[CAE-AN]{\includegraphics[width=0.33\textwidth]{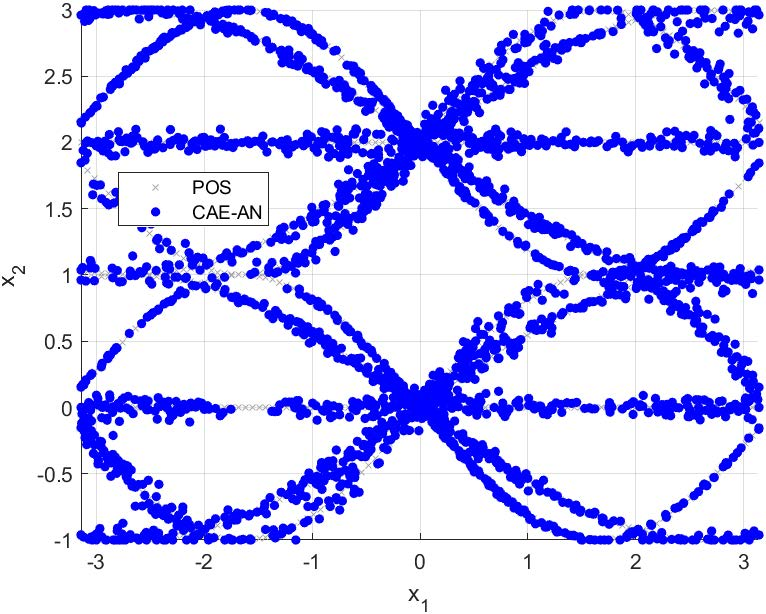}%
			\label{fig_first_case}}
		\hfill 
		\subfloat[D-NSGA-II]{\includegraphics[width=0.33\textwidth]{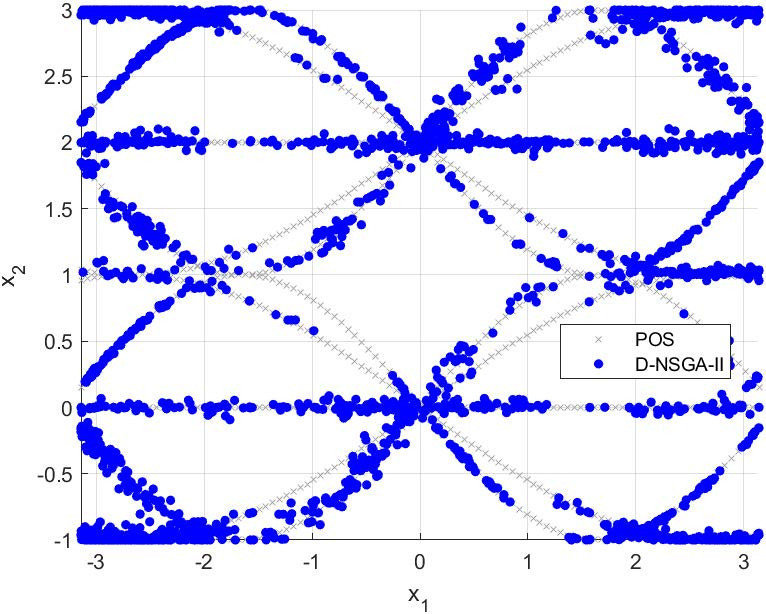}%
			\label{fig_second_case}}
		\hfill 
		\subfloat[DIP-DMOEA]{\includegraphics[width=0.33\textwidth]{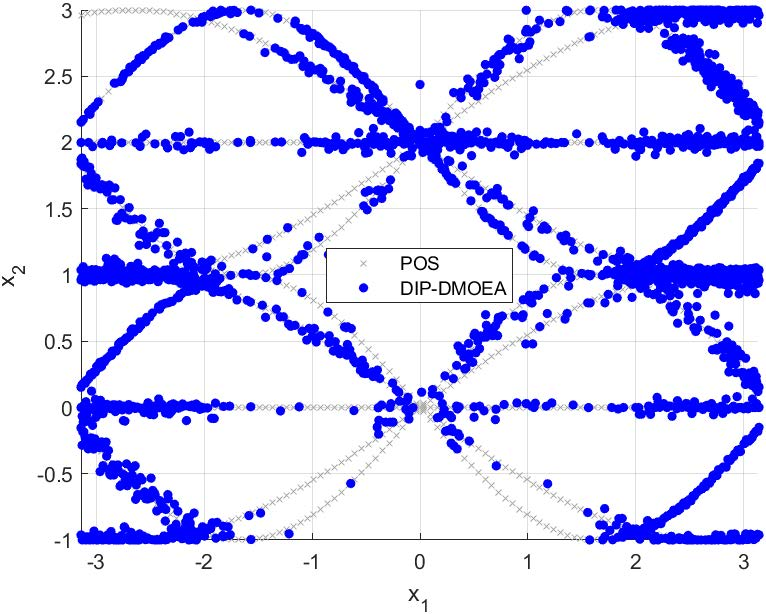}% 
			\label{fig_third_case}}
		\hfill 
		\subfloat[Tr-DMOEA]{\includegraphics[width=0.33\textwidth]{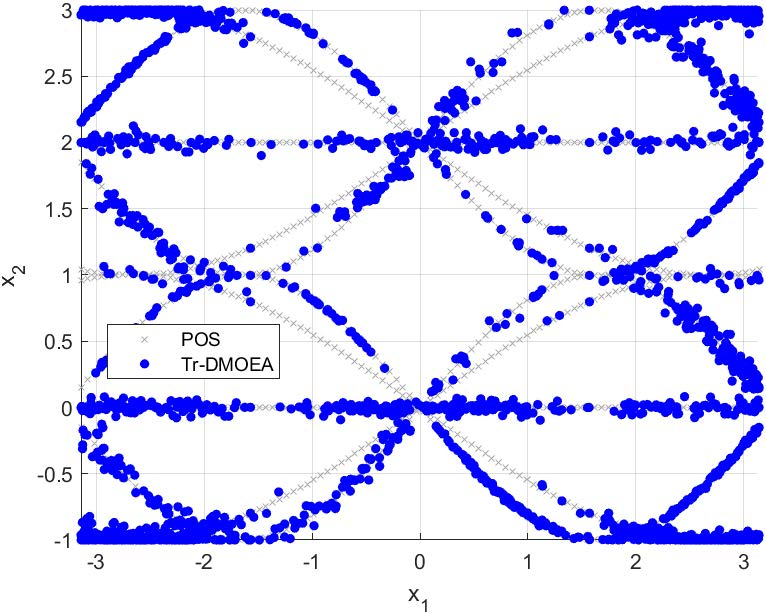}% 
			\label{fig_third_case}}
		\hfill
		\subfloat[D-DN-NSGA-II]{\includegraphics[width=0.33\textwidth]{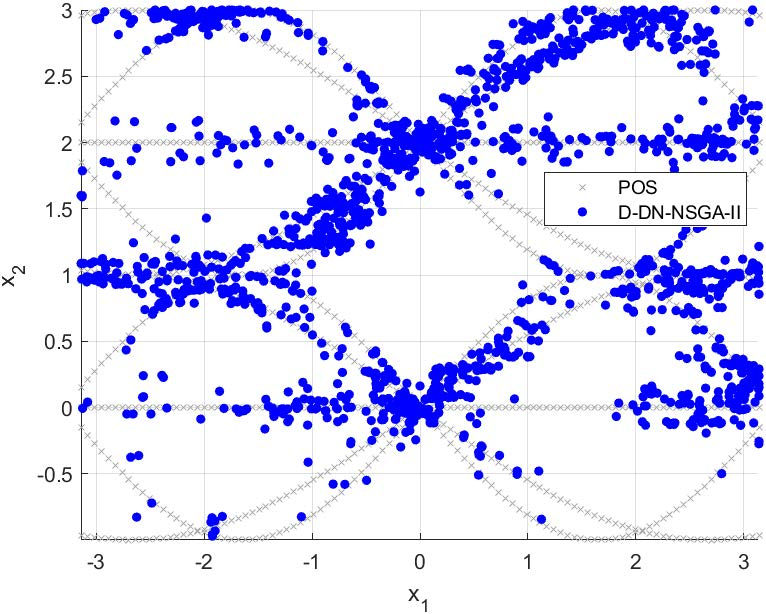}% 
			\label{fig_third_case}}
		\hfill
		\subfloat[D-MMEA-WI]{\includegraphics[width=0.33\textwidth]{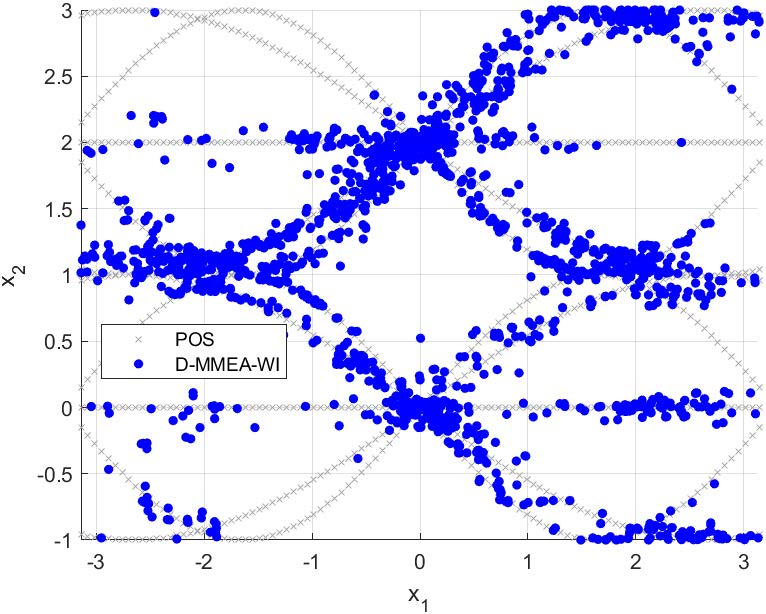}% 
			\label{fig_third_case}}
		\hfill
		\subfloat[D-MO-RING-SCD-PSO]{\includegraphics[width=0.33\textwidth]{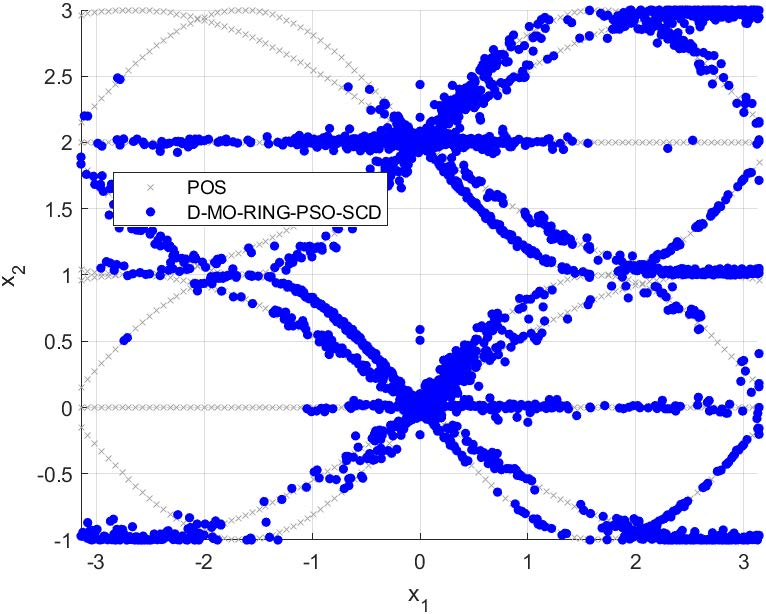}% 
			\label{fig_third_case}}
		\hfill
		\caption{Comparison of POS for Different Algorithms on the DMMF12 Test Instance}
		\label{fig_8} 
	\end{figure*}
	
	\subsection{Parameter Sensitivity Analysis}
	
	\begin{table}[!t]
		\centering
		\caption{Win Count Comparison for Parameter $\alpha$ across MIGD and MIGDx Metrics}
		\label{tab:alpha_wins_comparison}
		\begin{tabular}{cccc}
			\toprule
			\textbf{Parameter $\alpha$} & \textbf{MIGD Wins} & \textbf{MIGDx Wins} & \textbf{Total Wins} \\
			\midrule
			0.1 & 10 & 2 & \textbf{12} \\
			0.2 & 4 & 2 & \textbf{6}  \\
			0.3 & 7 & 2 & \textbf{9} \\
			0.4 & 3 & 3 & \textbf{6} \\
			0.5 & 19 & 31 & \textbf{50} \\
			0.6 & 3 & 4 & \textbf{7}  \\
			0.7 & 2 & 1 & \textbf{3}  \\
			0.8 & 1 & 2 & \textbf{3}  \\
			0.9 & 8 & 1 & \textbf{9} \\
			\bottomrule
		\end{tabular}
	\end{table}
	
	To investigate the influence of the decay factor $\alpha$ on the performance of the CAE-AN algorithm, a parameter sensitivity analysis was conducted. The factor $\alpha$ is central to the adaptive niche radius mechanism, dynamically adjusting the niche radius reduction rate based on evolutionary progress and local fitness variance. An inherent trade-off exists in the selection of $\alpha$, higher values accelerate convergence by shrinking niches more rapidly, which may compromise population diversity. Conversely, lower values help maintain diversity but can delay the algorithm's convergence to the POF.
	
	To determine the optimal value for $\alpha$, we systematically tested nine discrete values from the set $\{0.1, 0.2, \dots, 0.9\}$. For each $\alpha$ value, the CAE-AN algorithm was independently run 20 times on a standard test suite. As shown in Table~IV(with detailed data available in Table V and Table VI of the supplementary file), the experimental results clearly indicate that the algorithm's overall performance is optimal when $\alpha = 0.5$. Specifically, this setting not only achieved the lowest mean MIGD values on 11 out of the 12 test functions but also obtained the lowest mean MIGDx values on all 12 functions.
	
	In conclusion, a value of $\alpha = 0.5$ achieves an optimal balance between objective space convergence and decision space diversity for the CAE-AN algorithm. Therefore, this value is adopted as the default parameter setting for CAE-AN in all subsequent comparative experiments presented in this paper.

	\begin{table}[h]
		\centering
		\caption{STATISTICS OF PERFORMANCE COMPARISONS OF CAE-AN AND COMPARED VARIANTS ACCORDING TO MIGD AND MIGDx INDICATORS.}
		\label{tab:performance_comparison_vertical_2}
		
		\begin{tabular}{|l|c|}
			\hline
			\multicolumn{2}{|c|}{\textbf{MIGD}} \\
			\hline
			\textbf{Comparison} & \textbf{+/-/=} \\
			\hline
			CAE-AN\_none vs CAE-AN & 12/29/7 \\
			CAE-AN\_noC vs CAE-AN & 15/28/5 \\
			CAE-AN\_noAE vs CAE-AN & 11/35/2 \\
			CAE-AN\_noadaptive vs CAE-AN & 13/28/7 \\
			\hline
		\end{tabular}
		
		\vspace{1em} 
		
		\begin{tabular}{|l|c|}
			\hline
			\multicolumn{2}{|c|}{\textbf{MIGDx}} \\
			\hline
			\textbf{Comparison} & \textbf{+/-/=} \\
			\hline
			CAE-AN\_none vs CAE-AN & 15/30/3 \\
			CAE-AN\_noC vs CAE-AN & 9/36/3 \\
			CAE-AN\_noAE vs CAE-AN & 2/45/1 \\
			CAE-AN\_noadaptive vs CAE-AN & 15/32/1 \\
			\hline
		\end{tabular}
		
		\begin{flushleft}
			\small
			“+”, “-”, and “=” denote CAE-AN performs significantly worse than, better than, and similar to the compared algorithm based on the Wilcoxon’s rank sum test at 0.05 significant level.
		\end{flushleft}
	\end{table}
	
	\subsection{Ablation Study}
	To quantitatively evaluate the contributions of the dynamic response mechanism and the adaptive niching strategy, a series of ablation studies were conducted, comparing the performance of the complete CAE-AN algorithm against its four key ablated variants: the baseline method CAE-AN\_none, which removes both the prediction and adaptive niching core modules; CAE-AN\_noC, which omits the cluster matching component from the dynamic response; CAE-AN\_noAE, which replaces the AE-based prediction with a simple linear model; and CAE-AN\_noadaptive, which substitutes the adaptive strategy with a conventional niching technique. All algorithms were run on the DMMF test suite, and a Wilcoxon rank-sum test was employed to perform a significance analysis on the MIGD and MIGDx metrics. The statistical results summarized in Table~V (with detailed data available in Table III and Table IV of the supplementary file) provide evidence that the complete CAE-AN algorithm significantly outperforms all its ablated variants in terms of performance in both the objective and decision spaces. Further analysis of these variants reveals the distinct contributions of each core mechanism. For instance, the performance of CAE-AN\_noadaptive degrades significantly in the decision space, a failure rooted in its reliance on a conventional niching technique that lacks radius adaptation. This prevents it from effectively balancing the maintenance of population diversity with the enhancement of convergence pressure, making it prone to losing solutions when searching for multiple, disjoint POSs. Similarly, the performance of CAE-AN\_noC highlights the necessity of the cluster matching component, as its absence causes the prediction-only model to likely guide the population toward only a few regions. This results in an initial population for the static optimization phase that lacks diverse distributional information, thereby preventing the adaptive niching strategy from realizing its full potential and ultimately hindering convergence to the complete POS.

	\subsection{Real-world Application}
	To evaluate the practicality of the CAE-AN algorithm, this paper designed and solved a case study in dynamic multi-modal transportation planning. The domain of transportation planning was selected as the application subject due to its inherent dynamic features and multi-modal complexity, which involves diverse transportation modes with varying capacities and energy efficiencies, and where optimal solutions are often time-dependent.
	
	This paper utilizes a simplified model of a single-road segment to model the traffic flow allocation among three modes: cars, buses, and bicycles. A bi-objective optimization model was formulated with the goal of minimizing both total network travel time and total energy consumption. Congestion effects arising from interactions between cars and buses are characterized by the Bureau of Public Roads (BPR) function. Bicycle travel time is modeled primarily based on cycling speed, assuming they have dedicated rights-of-way.
	
	To solve this optimization problem, CAE-AN and five other comparison algorithms were applied. Consistent experimental parameters were maintained for all algorithms: a population size of 200 and 20 independent executions per algorithm. Algorithm performance was quantitatively evaluated using the Hypervolume (HV) indicator, a common standard for measuring the quality of multi-objective solution sets.
	
	The mean HV values for each algorithm are summarized in supplementary material. The results indicate that CAE-AN significantly outperforms all comparison algorithms on this transportation planning model. This superior performance underscores the significant potential of CAE-AN for addressing real-world DMMO tasks.
	\section{CONCLUSION}
	The field of DMMO faces critical challenge, particularly in the lack of comprehensive benchmarks and algorithms capable of effectively integrating dynamic adaptation with multimodal solution maintenance. To address these challenges, this paper presents two primary contributions. First, we have constructed a new benchmark suite, named DMMF, by synthesizing characteristics from existing dynamic and multimodal test problems. This suite comprises 12 distinct functions, each configurable with four environmental settings, designed to systematically encapsulate diverse dynamic and multimodal challenges, thereby providing a more rigorous evaluation platform for algorithm research.
	
	Secondly, this paper proposes a DMMOEA, named CAE-AN. The core innovation of the algorithm lies in its CAE strategy, which first identifies multiple optimal solution sets via a clustering algorithm and subsequently utilizes an AE model to predict the new positions of these sets when the environment changes. This mechanism is designed to effectively maintain the algorithm's convergence and diversity in dynamic environments. Furthermore, the algorithm integrates an adaptive niching strategy within its static optimizer to further enhance population diversity.
	
	Extensive comparative experiments were conducted, evaluating CAE-AN against six representative algorithms on the DMMF suite. Empirical results, quantified using MIGD, MIGDx, consistently demonstrated the superior convergence and diversity performance of CAE-AN. Sensitivity analysis further validated the robustness of the adaptive niching component, identifying $\alpha$=0.5 as an effective parameter setting.
	
	The field of DMMO remains emergent, necessitating continued research into both evaluation frameworks and algorithm design. The DMMF suite offers a contribution towards more comprehensive benchmarking. CAE-AN represents progress in synergizing predictive dynamic handling with multimodal search techniques. Although CAE-AN yielded strong results, opportunities exist for refining the interplay between its components. Future research should prioritize the development of unified performance metrics assessing both dynamic tracking fidelity and multimodal solution quality. Investigating advanced strategies for coupling environmental adaptation and multimodal diversity maintenance is also essential for creating algorithms that adeptly balance exploration and exploitation across complex, temporally evolving landscapes.

\end{document}